\documentclass[11pt,a4paper]{article}
\usepackage[hyperref]{acl2019}
\usepackage{times}
\usepackage{latexsym, microtype}

\usepackage{url}
\usepackage{tipa}
\usepackage{graphicx}
\usepackage{siunitx}
\usepackage{amssymb}
\usepackage{caption,subcaption,graphicx}
\usepackage{multirow}

\aclfinalcopy 
\def\aclpaperid{1268} 

\graphicspath{ {diagrams/} }

\title{Are we there yet? Encoder-decoder neural networks\\
as cognitive models of English past tense inflection}

\author{
Maria Corkery\\
{\small {\tt mcorkery@inf.ed.ac.uk}}
\And
Yevgen Matusevych \\
{\small {\tt yevgen.matusevych@ed.ac.uk}}
\And
Sharon Goldwater \\
{\small {\tt sgwater@inf.ed.ac.uk}}
\AND
  {\rm School of Informatics}\\
  University of Edinburgh\\
}

\date{}

\begin{document}
\maketitle
\begin{abstract}

The cognitive mechanisms needed to account for the English past tense have long been a subject of debate in linguistics and cognitive science.  Neural network models were proposed early on, but were shown to have clear flaws. Recently, however, \citet{kirovcotterell} showed that modern encoder-decoder (ED) models overcome many of these flaws. They also presented evidence that ED models demonstrate humanlike performance in a nonce-word task. Here, we look more closely at the behaviour of their model in this task. We find that (1) the model exhibits instability across multiple simulations in terms of its correlation with human data, and (2) even when results are aggregated across simulations (treating each simulation as an individual human participant), the fit to the human data is not strong---worse than an older rule-based model. These findings hold up through several alternative training regimes and evaluation measures. Although other neural architectures might do better, we conclude that there is still insufficient evidence to claim that neural nets are a good cognitive model for this task.

\end{abstract}

\section{Introduction}

For over 30 years, the English past tense has served as both inspiration and testbed for models of language acquisition and processing \citep[][etc.]{rumelhartmcclelland, pinkerprince, marcus1995, plunkettjuola, pinker2002, albrighthayes, seidenberg2014, kirovcotterell, blything2018}. One of the most well-known debates centres on whether the apparently rule-governed regular past tense is indeed represented cognitively using explicit rules. \citet{rumelhartmcclelland} famously argued against this hypothesis, presenting a neural network model intended to 
capture both regular and irregular verbs with no explicit rules.
However, \citet{pinkerprince} presented a scathing rebuttal, pointing out both theoretical and empirical failures of the model. In their alternative (dual-route) view, 
the regular past tense is categorical and captured via explicit rules, while irregular past tenses are memorized and can (occasionally) generalize via gradient analogical processes \cite{pinkerprince,prasada1993}. Their 
arguments were so influential
that although neural networks gained considerable traction in cognitive science more generally \cite{bechtel1991, mccloskey1991, elman96}, many linguists dismissed the whole approach.\footnote{Though see \citet{seidenberg2014}, who argue that some of the core ideas, such as the focus on statistical 
learning,
have nevertheless permeated the study of language.}

With the recent success of deep learning in NLP, however, there has been renewed interest in exploring the extent to which neural networks capture human behaviour in psycholinguistic tasks \cite[e.g.,][]{linzen2018,linzen2019}. In particular, \citeauthor{kirovcotterell} (\citeyear{kirovcotterell}; henceforth K\&C) revisited the past tense debate
and 
showed that modern sequence-based encoder-decoder (ED) models overcome many of the criticisms levelled at  \citeauthor{rumelhartmcclelland}'s model. Specifically, these models permit variable-length input and output that represent sequential ordering; can reach near-perfect accuracy on both regular and irregular verbs seen in training; and (using multi-task learning) can effectively generalize phonological rules across different inflections.

These primary claims are undoubtedly correct (and indeed, we replicate the accuracy results below). However, we take issue with another part of K\&C's work, in which they claim that their ED model also effectively models human behaviour in a nonce-word experiment (i.e., \emph{wug} test, described below). We explore the model's behaviour on this task in detail, and conclude that its ability to model humans is considerably weaker than K\&C suggest. 

In particular, we begin by showing that multiple simulations of the same model (with different random initializations) result in very different correlations with the human data. To ensure that this instability is not just due to the evaluation measure, we introduce an alternative measure, but still find unstable results. We then consider whether treating individual simulations as individual participants (rather than as a model of the average participant) captures the human data better. This aggregate model does show some high-level similarities to the human participants: both model and humans tend to produce irregulars more frequently for nonce words that are similar to many real irregular verbs. However, the model is still poor at capturing fine-grained distinctions at the level of individual verbs.
We conclude that, although deep learning approaches overcome many of the problems of earlier neural network models, there is still insufficient evidence to claim that they are good models of human morphological processing.

\section{Background}

\subsection{Nonce word experimental data}

Like K\&C, we use data from two experiments run by \citeauthor{albrighthayes} (\citeyear{albrighthayes}; henceforth A\&H). In Experiment 1, using a dialogue-based prompt, A\&H presented participants auditorily with nonce ``verbs'' that are phonotactically legal in English  (e.g., \textit{spling, dize}), and prompted participants to produce past tense forms of these verbs, resulting in a data set of \textbf{production probabilities} of various past tense forms.
In Experiment 2, participants first produced each past tense form (as in Experiment 1) and were then asked to rate the acceptability of either two or three possible past tense forms for that verb---one regular, and one or two potential irregulars. For example, for \emph{scride} \textipa{/skr"aId/}, participants rated \emph{scrided} \textipa{/skr"aId@d/} (regular), \emph{scrode} \textipa{/skr"oUd/} and \emph{scrid} \textipa{/skr"Id/} (irregular). This gives a data set of past tense form \textbf{ratings}.

Most of A\&H's own analyses rely on the ratings data, but the ED model is a model of production, so we follow K\&C and use the data from Experiment~1. The data is coded using the same set of suggested forms that were rated in Experiment~2: for each nonce word, A\&H counted how many participants produced the regular form, the irregular form (or each of the two forms, if there are two), and ``other'' (any other past tense form that was not among those rated in Experiment~2). The counts are normalized to compute production probabilities for each output form.

The nonce words used by A\&H were carefully chosen according to several criteria. First, they are phonologically ``bland'': i.e., not unusual-sounding as English words (as confirmed by a pre-test with participants). Second, as explained in the following section, they fall into several categories designed to test A\&H's hypothesis that (contra \citeauthor{prasada1993}, \citeyear{prasada1993}), \emph{both} regular and irregular past tense forms exhibit gradient (and not categorical) effects. 

\subsection{A\&H's model and islands of reliability}
\label{sec:ior}
To explain the categories of nonce words (which we will refer to in our analyses), we briefly describe A\&H's theory of past tense formation, which they implement as a computational model. 
The model postulates that speakers maintain a set of explicit structured rules that capture inflectional changes at different levels of generality. For example, a speaker might have rules such as:
\vspace{-1mm}
\begin{itemize}
  \setlength\itemsep{0.1em}
  \item /$\varnothing$/ $\rightarrow$ /\textipa{@d}/ if verb matches [X \{/d/, /t/\} \underline{\hspace{5mm}}]\\ 
  based on, e.g., \textit{want, need, start}.
  \item /i/ $\rightarrow$ /\textipa{E}/ if verb matches [X \{/r/, /l/\} \underline{\hspace{5mm}} /d/]\\
  based on, e.g., \textit{read, lead, breed}. 
\end{itemize}
\vspace{-1mm}
where X represents arbitrary phonological material and \underline{\hspace{5mm}} is the location of the changing material.
Each rule is given a confidence score based on its precision and statistical strength (the number of cases to which it could potentially apply). When a nonce word is presented, several rules may apply (e.g., the two rules above for \emph{gleed}), and the goodness of each possible past tense is determined by the confidence score of the corresponding rule. 

Crucially, A\&H's model can learn multiple rules that all produce regular past tense forms, but with phonological contexts of different specificity, hence different confidence scores. Therefore, some nonce words may reside in so-called ``islands of reliability'' (IOR) for regular verbs: that is, there is an applicable regular rule that has a very high confidence score. Meanwhile other nonce words might also be considered regular, but with lower confidence. Thus, the model predicts gradient effects even for regular inflection. It also predicts gradient effects for irregular inflection, since there can be IORs for irregular rules as well.

To test these predictions, A\&H chose four types of nonce words: those residing in an IOR for regulars, for both regulars and irregulars, for irregulars only, or for neither. They also included several nonce verbs similar to \emph{burn--burnt, spell--spelt}, and some that might potentially elicit single-form analogies. Their results (discussed further in Section~\ref{sec:experiments}) showed that the different IOR categories were indeed treated differently by participants.

\subsection{Evaluating models}
\label{sec:bgeval}

To go beyond coarse-grained analysis based on the IOR categories, both A\&H and K\&C evaluate their models by correlating model output with the human data at the level of individual past tense forms. 
Correlations are computed between the human data (either production probabilities or ratings) and the model scores for each form. The regulars and irregulars are treated separately. 
That is, the irregular correlation value is computed by considering the average human production probability (or rating) for each suggested irregular past tense, and comparing these with the model scores for those same forms. The correlation for regulars is computed analogously. Regulars and irregulars are treated separately because the scores for regulars are nearly always larger, so if all forms were considered at once, a baseline that simply assigned (say) 1 to regulars and 0 to irregulars would already achieve a high correlation with humans.

We initially follow K\&C in computing the Spearman (rank) correlation against the production probabilities, and later also examine Pearson (linear) correlations and ratings data.

\section{Methods}
\label{methods}

\subsection{Model and hyperparameters}

We adopt the encoder-decoder architecture used by K\&C, as well as their implementation framework and hyperparameters. Encoder-decoder models are a type of recurrent neural network (RNN) introduced for machine translation \cite{sutskever2014} but also often used for other sequence-to-sequence transductions, such as morphological inflection and lemmatization \cite{kann2016,bergmanis2018}. The encoder is an RNN that reads in the input sequence (here, a sequence of characters representing the phonemes in the present tense verb form) and creates a fixed-size vector representation of it. The decoder is another RNN that takes this vector as input and decodes it sequentially, outputting one symbol at each  timestep (here, the phonemes of the past tense form). 
The ED model with attention \citep{bahdanau} is implemented in OpenNMT \cite{opennmt}.\footnote{In early tests, we also tried the Nematus toolkit with hyperparameters following \cite{kann2016,bergmanis2018}; the pattern of results was similar.} 
It has two bidirectional LSTM encoder layers and two LSTM decoder layers, 300-dimensional character embeddings in the encoder, and 100-dimensional hidden layers in the encoder and decoder. The Adadelta optimizer \cite{zeiler} is used for training, with the default beam size of 12 for decoding. The batch size is 20, and dropout is applied between layers with a probability of $0.3$. Except where otherwise noted below, all models were trained for 100 epochs. 

\subsection{Training data}

To compare our results to both A\&H and K\&C, we use their corresponding training sets, both based on data from CELEX \cite{celex}. 
A\&H's training data contains
all verbs listed in CELEX with a lemma frequency of 10 or more (4253 verbs, 218 of which are irregular).
We use A\&H's American English IPA phonemic transcriptions, to match the nonce word experiment (which was carried out with American English speakers), and also follow them in using the nonce words as the unseen test set rather than creating dev/test splits from the CELEX data.
As argued by A\&H, adult English speakers will have been exposed to all of the real verbs many times and would be able to correctly produce the past tense of all of them. Adults' generalization to nonce words is therefore predicated on their knowledge of this entire training set (including, crucially, all of the irregular forms).

For our second training set, we obtained the data from K\&C, which is a subset of A\&H's: it contains 4039 verbs, 168 of which are irregular---that is, 50 real irregular verbs are missing. 
Examples of verbs that are missing from the K\&C data include \emph{do--did} and \emph{use--used}. K\&C also  randomly divided their data into training, development, and test sets, but we weren't able to obtain these splits, so (since we are using the nonce words for test data) we simply use all 4039 verbs as training data.
We include results using the K\&C's data mainly to allow closer (though still not exact) comparison with their work, but we feel that A\&H's training data, which includes all the irregulars, more accurately reflects adult linguistic exposure. 

\begin{table}
  \centering
   \begin{tabular}{llS[table-format=<1.4]}
      \hline
      Rank      & nold \textipa{/n"oUld/}                & \text{Probability}       \\
      \hline
      1         & \textbf{nolded} \textipa{/n"oUld@d/}   & .9869            \\
      2         & nelt \textipa{/n"Elt/}                 & .0120            \\
      3         & neelded \textipa{/n"i:ld@d/}           & .0004            \\
      4         & nelded \textipa{/n"Eld@d/}             & .0004            \\
      5         & \textbf{neld} \textipa{/n"Eld/}        & .0001            \\
      \hline
      Rank      & murn \textipa{/m"@rn/}                 &  \text{Probability}      \\
      \hline
      1         & \textbf{murned} \textipa{/m"@rnd/}     & .8636            \\
      2         & \textbf{murnt} \textipa{/m"@rnt/}      & .1363            \\
      3         & murn \textipa{/m"@rn/}                 & < .0001 \\
      4         & murnaid \textipa{/m"@rneId/}           & < .0001 \\
      5         & murnoo \textipa{/m"@rnu:/}             & < .0001 \\
      \hline
    \end{tabular}
    \caption{Top 5 outputs from two sample beams, for the nonce words \emph{nold} and \emph{murn}. Past tenses suggested by A\&H are bolded.
    For \emph{nold}, one suggested past tense form, \emph{nold} \textipa{/n"oUld/}, is missing from the top 5.
        }
    \label{table:samplebeam}
\end{table}

It has been argued that morphological generalization in humans is governed by type frequencies rather than token frequencies \citep{bybee1997, pierrehumbert2001}. Modelling evidence, including from A\&H, also supports the idea that token frequencies are ignored or severely downweighted (i.e., effectively using log frequencies: \citealp{odonnell2015,goldwater2006}).
We therefore follow A\&H and K\&C in training our models on the list of distinct word types, with each type occurring once in the training data.

\subsection{Evaluation}
\label{s:evalmeasures}
We report three different evaluation measures. First, we compute {\bf training set accuracy}: the percentage of verbs in the training data for which the model's top-ranked output is the correct past tense form. This is largely a sanity check and test of convergence: a fully-trained model of adult performance should have near-perfect training set accuracy.

Next, as described in Section~\ref{sec:bgeval}, we report Spearman's rank {\bf correlation ($\mathbf{\rho}$)} of the model's probabilities for the various nonce past tense forms with the human production probabilities.
The probability for each suggested past tense form was obtained by forcing the model to output that form (e.g., providing \emph{scride} as input and forcing it to output \emph{scrid}).
This made it possible to get probabilities for forms that did not occur in the beam (the list of most likely forms output by the model).

Finally, we introduce a third measure, motivated further in Section~\ref{sec:exp1results}, {\bf complete recall@5}:
\begin{equation}
    \text{CR@5} = \frac{1}{n} \times \sum_{i=1}^{n}{\left[S_i \subseteq B_i \right]}
\end{equation}
where $n$ is the total number of nonce verbs, $S_i$ is the set of A\&H's suggested past tense forms for verb $i$, $B_i$ is the set of the top five verbs in the model's beam for $i$, and $\left[S_i \subseteq B_i \right] = 1$ if all verbs from $S_i$ appear in $B_i$, and $0$ otherwise.
For example, 
a model which only processed the two verbs in Table~\ref{table:samplebeam} would have a CR@5 of 0.5, since the beam includes all suggested past tenses for \emph{murn} (\emph{murned, murnt}), but not for \emph{nold} (\emph{nolded, nold, neld}).\footnote{Not all of A\&H's suggested forms were actually produced by participants, but all of them seem plausible and we felt that a good model should rank them higher than most other potential past tenses, i.e., they should be included within a small beam size. Indeed, in cases where they are not (e.g., \emph{nold} in Table~\ref{table:samplebeam}) we do typically see much less plausible forms (such as \emph{neelded}) included in the beam.
}

\section{Experiments}
\label{sec:experiments}

\subsection{Experiment 1: Model variability}
\label{exp1}
\label{exp1_methods}

Our first experiment aims to replicate K\&C's results showing that (a) the model is able to produce the past tense forms of training verbs with near-perfect accuracy, and (b) its correlation with human data on the nonce verb test set is higher than that of A\&H's model.
In K\&C's paper, these results were based on a 
single trained model. 
Here we trained 20 models (10 on each training set) initialized with different random seeds.

\begin{table}
    \setlength{\tabcolsep}{4pt}
    \centering
    \begin{tabular}{llll}
        \hline
        Data & all & regular & irregular \\
        \hline
        K\&C & 99.79 (0.05) & 99.92 (0.04) & 96.90 (1.06) \\
        A\&H & 99.51 (0.04) & 99.86 (0.07) & 92.98 (1.18) \\
        \hline
    \end{tabular}
    \caption{Mean training set accuracy (in \%, with standard deviations in brackets), averaged over $10$ runs for each training set with different random seeds. Oracle accuracy is $99.85$\%  on the K\&C data and $99.55$\% on the A\&H data, due to homophones and forms with multiple past tenses. In order to do better on irregulars, the model would have to get more of the regulars wrong. 
    }
    \label{table:accuracies}
\end{table}

\label{sec:exp1results}

\begin{figure}
  \centering
  \includegraphics[width=\linewidth]{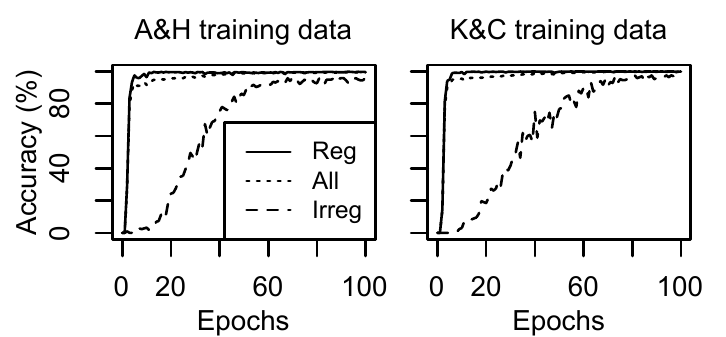}
  \caption{Accuracy values on the training set during training for one model per training set.}  
  \label{fig:convergence}
\end{figure}

\paragraph{Accuracy} Table \ref{table:accuracies} lists the mean and standard deviation of training set accuracy for each of the two training sets. It is not possible to get 100\% accuracy because the training sets contain some homophones with different past tenses (e.g., \emph{write--wrote} and \emph{right--righted}), and some verbs which have two possible past tenses (e.g., \emph{spring--sprung} and \emph{spring--sprang}). Nevertheless, the models get very close to the best possible accuracy,
confirming K\&C's finding that they learn both regular and irregular past tenses of previously seen words within 100 epochs.
Example convergence plots are shown in Figure~\ref{fig:convergence}, illustrating that the models learn regular verbs very quickly, and irregular verbs more slowly, but both are learned well after 60--80 epochs.

\begin{figure}
  \centering
  \includegraphics[width=\linewidth]{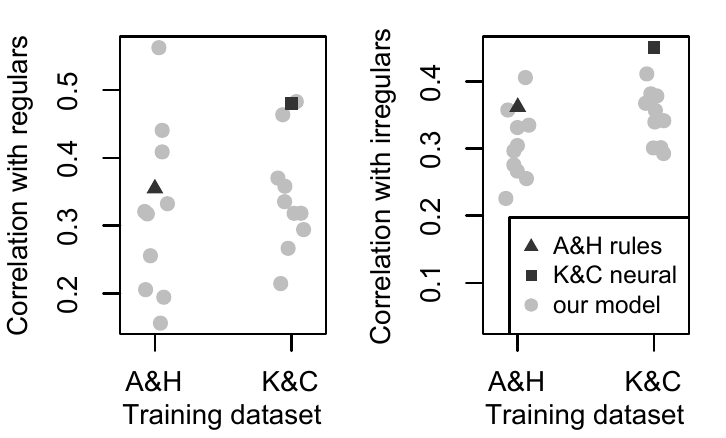}
  \caption{Spearman correlation coefficients between model scores and human production probabilities, using the A\&H and K\&C training data. Values reported by K\&C and A\&H are shown in addition to those of our models. Horizontal jitter is added for readability.}
  \label{fig:pps_corrs_sep}
\end{figure}

\paragraph{Correlation} Despite having consistently high accuracy on real words, Figure \ref{fig:pps_corrs_sep} shows that models with different random initializations vary considerably in their correlation with human speakers' production probabilities on nonce words, from
$0.15$ to $0.56$ for regulars, and from $0.23$ to $0.41$ for irregulars. 
K\&C's reported results  
are at the high end of what we obtained, suggesting that they are likely not representative.
 
On the other hand, we were concerned that the variability in the correlation measure might be due to an artefact: the vast majority of the beams returned by the model assign very high probability ($>98\%$) to the top item and little mass to anything else (as in the first example in Table~\ref{table:samplebeam}).\footnote{The skewedness of the beams is likely because  of the 
training/testing scenario, where
the model is effectively asked to do different tasks: at training time, it is trained to produce one correct past tense, while at test time, it's expected to produce a probability distribution over potential nonce past tenses. We could surely produce better matches to the human probability distributions by training directly to do so, but that wouldn't make sense as a cognitive model, since human learners are exposed only to correct past tenses, not to distributions.
}
Since the correlation measure is computed across different nonce forms, tiny changes in the beam probabilities of one nonce verb could change the ranking of (say) its regular past with respect to the regular past of another nonce word, even if the relative ranking of forms within each nonce's beam stayed the same. 

\begin{figure}
  \centering
  \includegraphics[width=\linewidth]{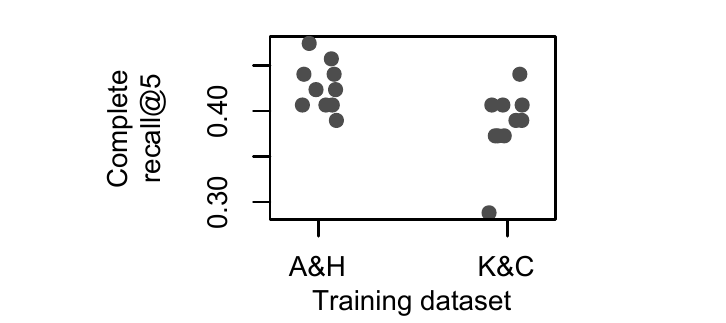}
  \caption{Complete recall@5
  for 20 models with different random seeds (10 with each training dataset). Horizontal jitter is added for readability.}
  \label{fig:complete_recall}
\end{figure}

\paragraph{CR@5 and second best forms}  The above observation motivated the CR@5 measure (Section~\ref{s:evalmeasures}). Rather than measuring the relative probabilities of past forms across different verbs, CR@5 considers the relative rankings of different past forms for each verb. However, CR@5 also yielded unstable results: 39--47\% on A\&H's data, and 29--44\% on K\&C's data, as shown in Figure \ref{fig:complete_recall}.

\begin{figure}
  \centering
  \includegraphics[width=\linewidth]{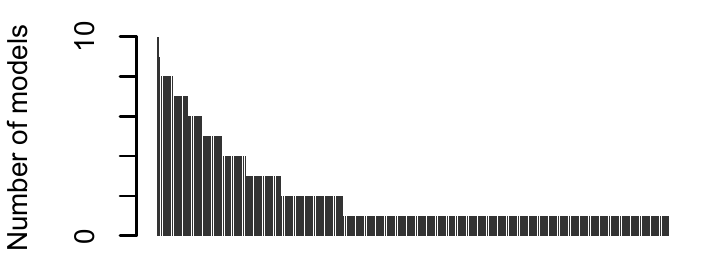}
  \caption{The number of models (of the 10 trained on the A\&H dataset) which agree on the second-place past tense form. The X-axis shows 281 different past tense forms (for 59 nonce words in the present tense), and the Y-axis shows, for each form, how many times a model places it in the second position in the beam.}
  \label{fig:secondplace}
  \vspace{-3mm}
\end{figure}

As a final exploration of the models' instability across different simulations, we looked at how often the models agree with each other on the verb occupying the first and the second position in the beam. While there is very high agreement on the most likely form (top of the beam) across the simulations---usually a regular past tense---very few forms in the second position are the same across simulations (see Figure \ref{fig:secondplace}).

\begin{figure*}
  \centering
  \includegraphics[width=1.0\linewidth]{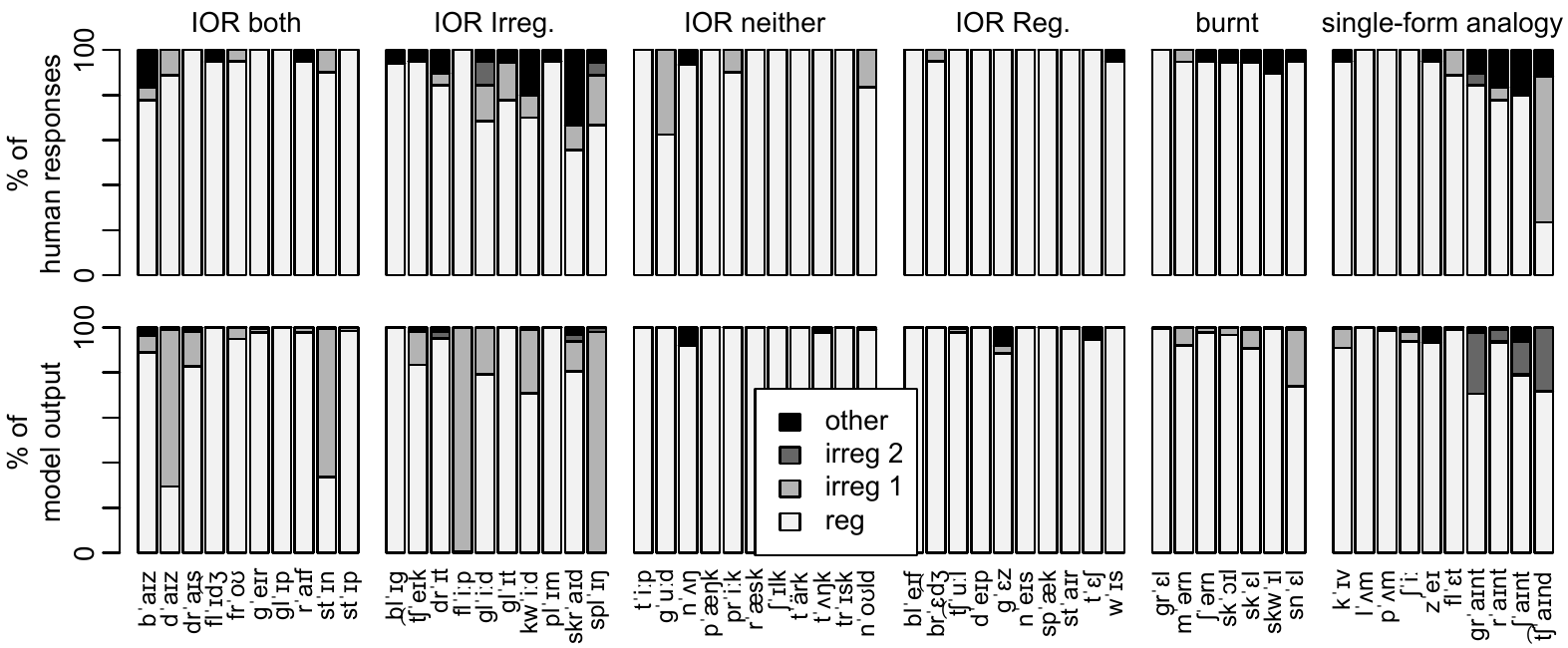}
  \caption{Percentage of regular, irregular, and ``other'' responses produced by humans (top) and the model (bottom).
  Each of the six blocks corresponds to a different category of nonce words (see Section~\ref{sec:ior}).
  }
  \label{fig:prod_simulation}
\end{figure*}

\paragraph{Summary} To recap, we find similar training set accuracy to what K\&C reported, but
the correlation scores between the model and the human data are generally lower, and the model exhibits unstable behaviour across different simulations. However, the unstable behaviour can potentially be accounted for, if each simulation is interpreted as an individual participant rather than as a model of the average behaviour of all participants.
In that case, we should aggregate results from multiple simulations in order to compare them to the human results, since production probabilities from A\&H's experiment were obtained by aggregating data over multiple participants.
The next experiment examines this alternative interpretation.

\subsection{Experiment 2: Aggregate model}

To simulate A\&H's production experiment with each simulation as one participant, we trained $50$ individual models on the A\&H training data\footnote{In the absence of clear differences between the model's performance on A\&H's vs.\ K\&C's data in Experiment 1, we only use  the more complete A\&H dataset henceforth.} using the same architecture and hyperparameters as before. We then sampled $100$ past tense forms for each verb from each model's output probability distribution. Each of the 5000 output forms (100 each from 50 simulated participants) was categorized either as (a) the verb's regular past tense form, \mbox{(b--c)} the first or second irregular past tense form suggested by A\&H, or (d) any other possible form.

For the aggregate model, the correlation measure is the only evaluation that makes sense. 
For regulars, correlation with
the human production probabilities was higher than in the previous experiment ($0.45$ vs.\ an average of $0.28$ in Experiment~1), but for irregulars it was lower ($0.19$ vs.\ $0.22$ in Experiment 1).  
The differences between the humans and aggregate model are clear from Figure \ref{fig:prod_simulation}, which shows the distribution of various past tense forms for both model and humans.  For example, in only one case did the humans produce an irregular more frequently than the regular (no-change past \emph{chind} for present \emph{chind}), whereas there are several cases where the aggregated model does so. Moreover, for the word \emph{chind} itself, the model prefers \emph{chound} rather than \emph{chind}.

In the previous experiment, we saw that individual models often rank implausible past tenses higher than plausible ones. 
However, we see here that on aggregate 
nearly all the model's proposed past tenses are those suggested by A\&H. Apparently, the unstable beam rankings wash out the implausible forms, i.e., the plausible forms on average occur nearer the top of the beam than any particular implausible form. In fact, the model actually produces fewer ``other'' forms than the humans. 

\begin{figure}
  \centering
\vspace{-5pt}  \includegraphics[width=\linewidth]{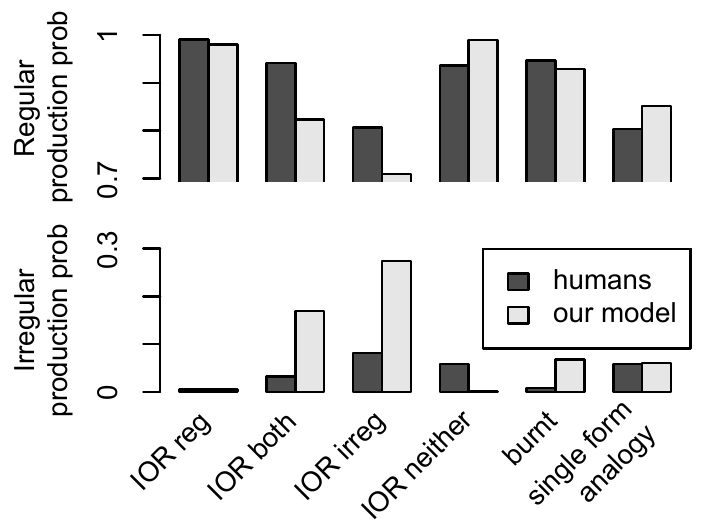}
\vspace{-20pt}
\caption{Mean production probabilities for regulars (top) and A\&H's suggested irregulars (bottom) in each of A\&H's categories of nonce words, for humans and for the aggregated ED model. }
  \label{fig:ior}
\end{figure}

We also looked at the model's average production of regular and suggested irregular forms for each of the six categories in Figure \ref{fig:prod_simulation}. The results, shown in Figure~\ref{fig:ior}, indicate that the model does capture the main trends seen in humans across these categories, but overall it is more likely to produce irregular forms. Together with the low overall correlation to human results and obvious differences at the fine-grained level, these results suggest that there are serious weaknesses in the ED model, even when results are aggregated across simulations. 

\section{Further analyses} 


\subsection{Is the model overfitting?}
\label{sec:overfitting}
We began by assuming that models should be trained at least until they achieve perfect performance on the training set, but perhaps 100 epochs is too much, and the model is just overfitting.
Training for less time might produce less skewed beam probabilities, more stable beam rankings, and perhaps better correlations with the human data.

To investigate this possibility, we took the 10 models originally trained on the A\&H dataset and computed the correlation with human data for regulars and irregulars after every 10 epochs of training. 
The highest correlation is achieved after only 10 epochs (0.47 for regulars and 0.50 for irregulars) and the beam probabilities are indeed less skewed: the average probability of the top ranked output is 0.92 after 10 epochs, vs.\ 0.97 after 100 epochs. However, the models average only 6.5\% accuracy on the real irregular words after 10 epochs, so it is difficult to argue that these are good models of human behaviour.\footnote{Early exposure to more irregulars could help in principle, so we also tried training the models on token or log token frequencies rather than type frequencies, but the resulting models' correlations with production probabilities were no higher than models trained on type frequencies (the same for log tokens, and lower for tokens).} 
It seems that the ED model displays a fundamental tension between correctly modelling humans on real words and nonce words.

\begin{table}
\begin{tabular}{llllll}
\hline
Data                                                                        & Cor.                       & Verbs  & A\&H          & Individ.    & Agg.         \\ \hline
\multirow{4}{*}{\begin{tabular}[c]{@{}l@{}}Pro-\\ duc-\\ tion\end{tabular}} & \multirow{2}{*}{$\rho$}     & reg.   & .35          & .32 (.12) & \textbf{.45} \\
                                                                            &                             & irreg. & \textbf{.36} & .31 (.05) & .19          \\ \cline{2-6} 
                                                                            & \multirow{2}{*}{\textit{r}} & reg.   & \textbf{.62} & .16 (.09) & .30          \\
                                                                            &                             & irreg. & .14          & .16 (.03) & \textbf{.17} \\ \hline
\multirow{4}{*}{\begin{tabular}[c]{@{}l@{}}Rat-\\ ing\end{tabular}}         & \multirow{2}{*}{$\rho$}     & reg.   & \textbf{.55} & .32 (.09) & .43          \\
                                                                            &                             & irreg. & \textbf{.57} & .39 (.08) & .31          \\ \cline{2-6} 
                                                                            & \multirow{2}{*}{\textit{r}} & reg.   & \textbf{.71} & .34 (.07) & .40          \\
                                                                            &                             & irreg. & \textbf{.48} & .35 (.06) & .40          \\ \hline
\end{tabular}
\caption{Correlations (using Spearman's $\rho$ and Pearson's $r$) between the models' output probabilities vs.\ human production probabilities and rating data. The data for the individual model is an average over 10 simulations (standard deviation shown in brackets). Highest correlation in each line is shown in bold.}
\label{table:allcorrs1}
\end{table}

\subsection{Rating data and correlations}
\label{sec:correlations}

We have so far evaluated all models against human production data. However, the A\&H model outputs unnormalized scores, so arguably it makes more sense as a model of ratings. A\&H also originally evaluated it using Pearson correlation. For completeness we report in Table~\ref{table:allcorrs1} the correlations for all models on both ratings data and production data, using both Spearman and Pearson coefficients. We find that the A\&H model does score better against ratings data, although surprisingly the ED models do too. More importantly, though, the A\&H model fits the human data best on 6 out of 8 measures.

\subsection{What is the model learning?}
\label{sec:representations}

\begin{figure*}
  \begin{minipage}{0.74\linewidth}
    \begin{subfigure}[t]{\linewidth}
      \centering
      \includegraphics[width=\linewidth]{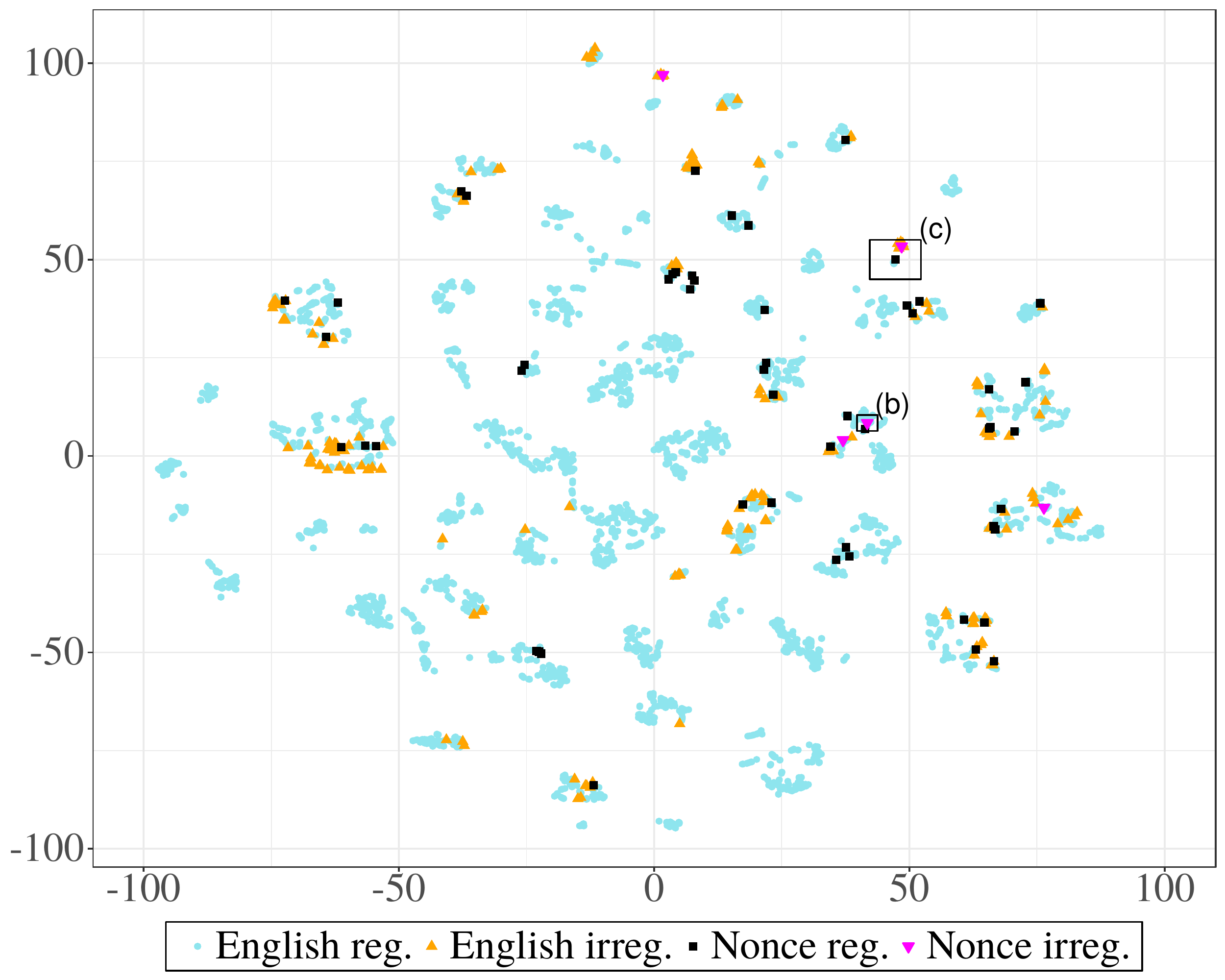}
      \caption{All verbs.}
    \end{subfigure}
  \end{minipage}
  \begin{minipage}{0.25\linewidth}
      \begin{subfigure}[t]{\linewidth}
      \vspace{-1pt}
        \includegraphics[width=\linewidth]{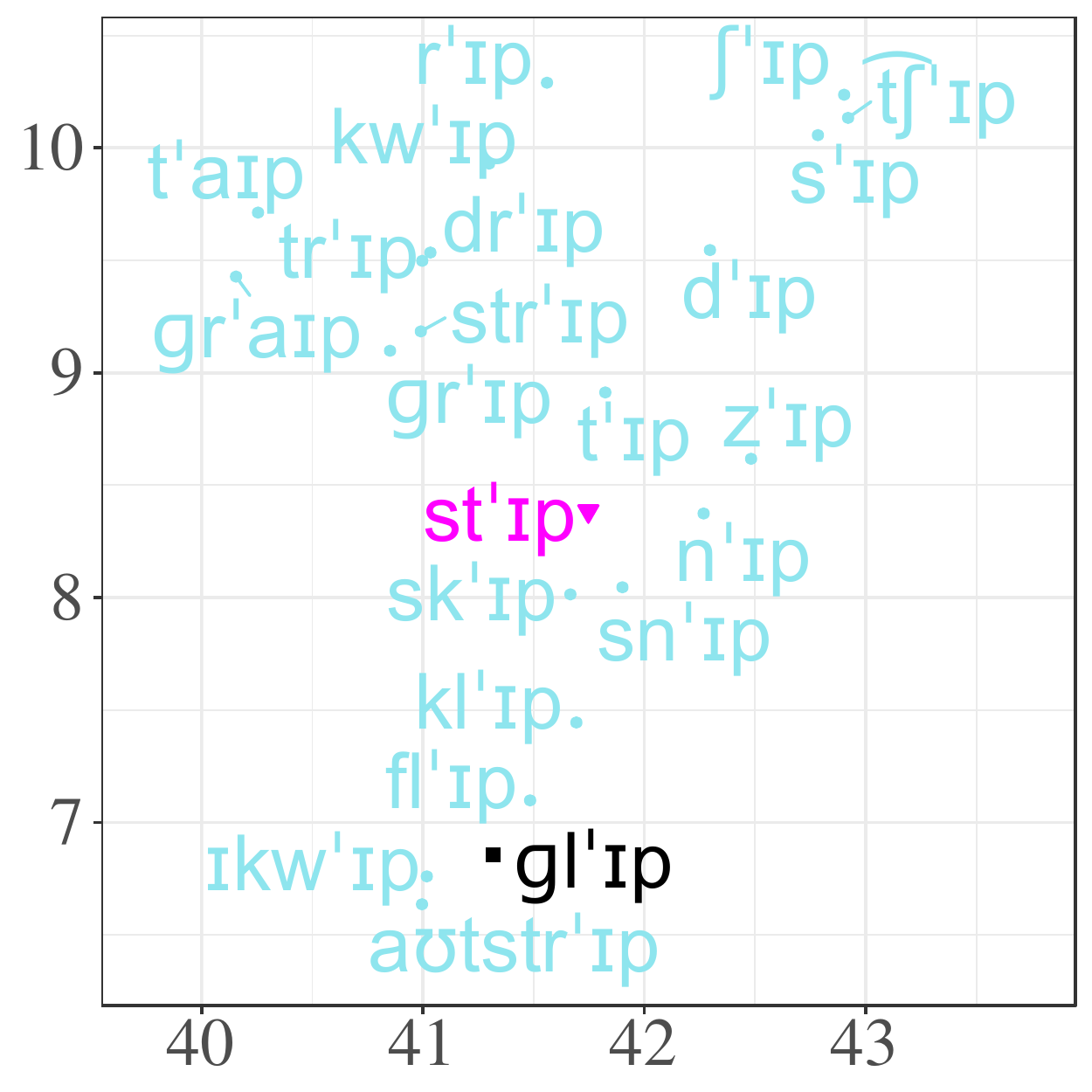}
        \caption{Zooming in on \textipa{/st"Ip/}.}
      \end{subfigure}
      \begin{subfigure}[t]{\linewidth}
      \vspace{22pt}
        \includegraphics[width=\linewidth]{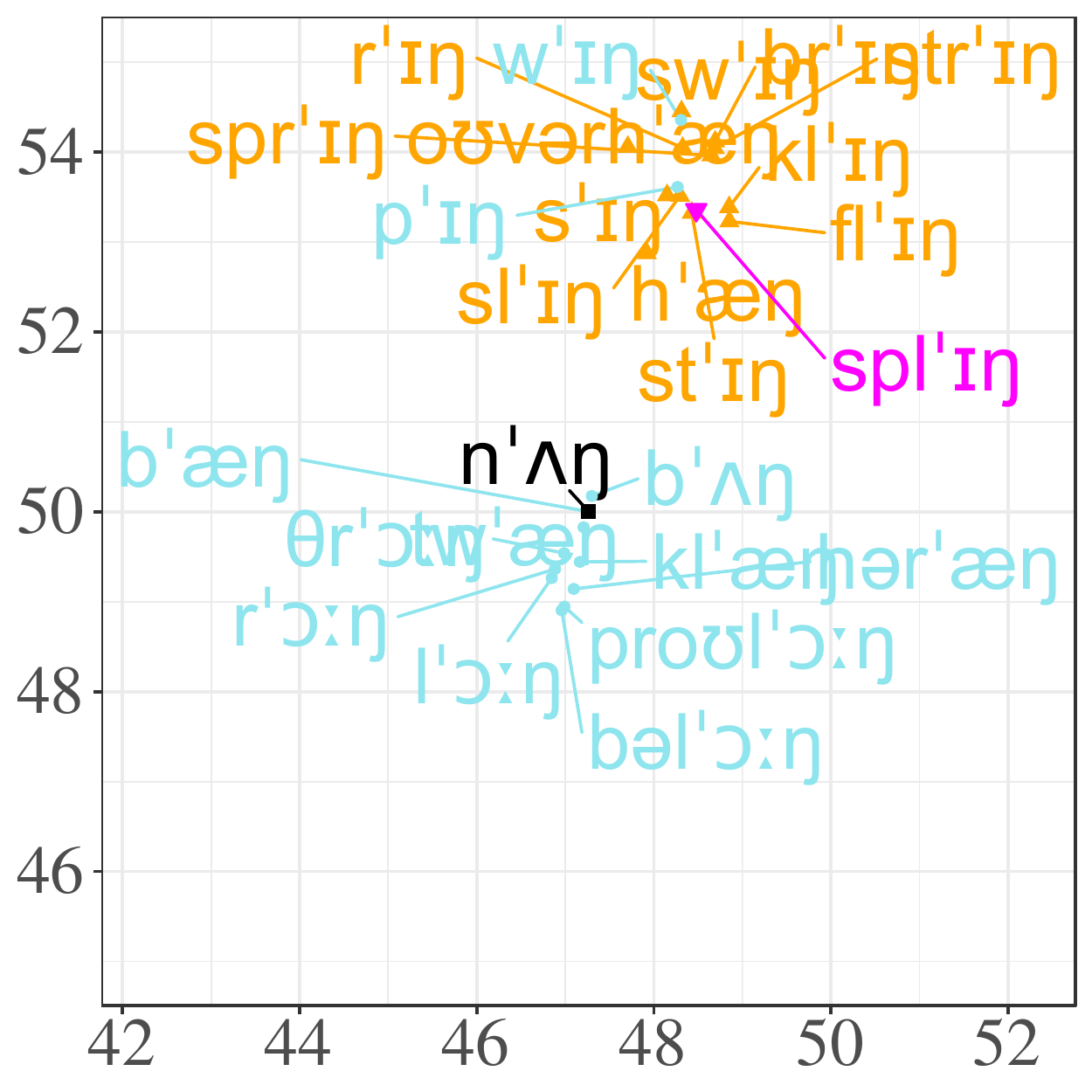}
        \caption{Zooming in on \textipa{/n"2N/}.}
      \end{subfigure}
  \end{minipage}
  \caption{A t-SNE plot of encoder state vectors for regular and irregular verb forms. (a) shows an an overview of all (real and nonce) verbs, and (b) and (c) zoom in on the boxed areas in (a).}
  \label{fig:tsne}
\end{figure*}

To examine the representations acquired by the model, we extract vectors from the encoder's hidden state. As the encoder is a bidirectional LSTM, we concatenate the two states at the last time step (after training on the A\&H data). Figure~\ref{fig:tsne}a shows a \mbox{t-SNE} visualization of hidden state vectors for both real and nonce verbs in one of our simulations. The model clearly groups the verbs into small clusters, and Figures~\ref{fig:tsne}b--c show that this clustering is based on the verbs' trailing phonemes, including some structure withing the clusters: e.g., \textit{strip} \textipa{/str"Ip/}, \textit{grip} \textipa{/gr"Ip/}, and \textit{trip} \textipa{/tr"Ip/} are next to each other in Figure~\ref{fig:tsne}b, and so are \textit{clip} \textipa{/kl"Ip/}, \textit{flip} \textipa{/fl"Ip/}, and \textit{glip} \textipa{/gl"Ip/}. It is not so clear, however, how the model decides on whether to produce a regular or an irregular form for nonce verbs. We do see some evidence in Figure~\ref{fig:tsne}c that nonce verbs similar to regular English verbs yield a regular form (note the regular neighbours of \textit{nung} \textipa{/n"2N/)}, and the same holds for irregulars (note the irregular forms around \textit{spling} \textipa{/spl"IN/}, for which the model produced \textit{splung}). However, the model also produces an irregular form (\textit{stup} \textipa{/st"2p/}) for \textit{stip} \textipa{/st"Ip/}, which is clearly surrounded by regular English verbs in Figure~\ref{fig:tsne}b.

We also tested whether the clustering by trailing phonemes is simply an artefact, by training another model on data where we reversed the order of the input phonemes in all cases
(e.g., \textipa{/w"IS/}--\textipa{/w"ISt/} [\emph{wish--wished}] becomes \textipa{/SI"w/}--\textipa{/tSI"w/}). This time, verbs were grouped based on their \textit{leading} phonemes---that is, the endings of the original verbs---suggesting that the model finds the regularities in the data regardless of the order of phonemes.

\begin{figure}
  \centering
  \includegraphics[width=\linewidth]{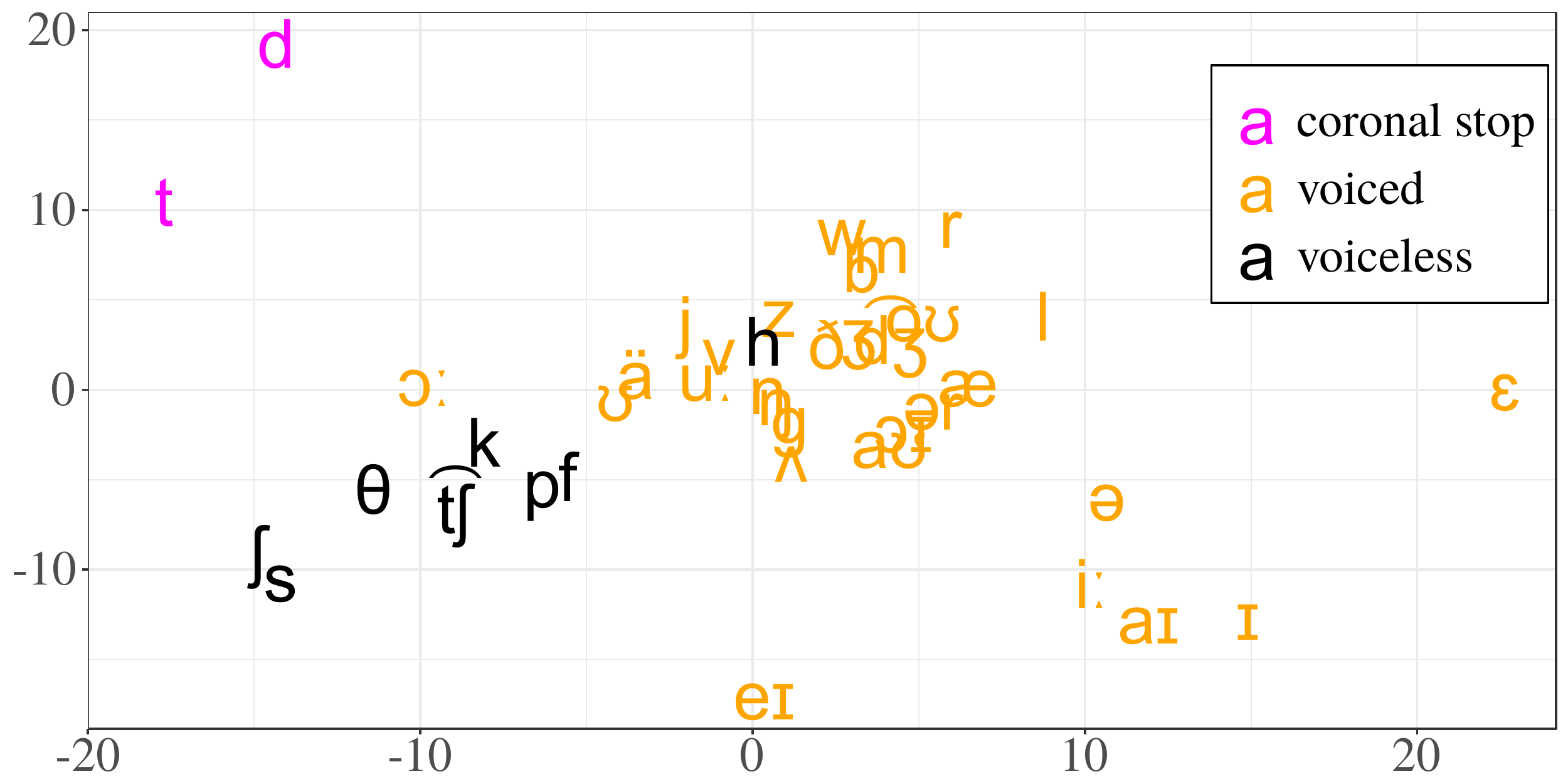}
  \caption{PCA plot of character-level (phoneme) vectors extracted from the decoder's hidden state. The phonemes are coloured based on the three different regular past-tense suffixes they would be followed by.}  \label{fig:pca}
\end{figure}

Finally, we investigated the model's phoneme representations, expecting a clustering corresponding to the three types of phonemes that trigger different endings in regular past tense forms: \textipa{/-Id/} after coronal stops \textipa{/t/} and \textipa{/d/}, \textipa{/-d/} after voiced consonants and vowels, and \textipa{/-t/} after voiceless consonants. We extract character-level vectors from the decoder hidden state, apply PCA (which worked better than t-SNE in this case) and visualize the resulting vectors. Figure~\ref{fig:pca} shows that the expected pattern has emerged
(except for /h/ in the `voiced' cluster, but this phoneme never appears at the end of English words).

 \section{General discussion and conclusions}

Our results confirm that, unlike earlier neural net models, the ED model has no trouble learning the past tense forms of verbs it is trained on. We found, however, that its behaviour on nonce verbs does not correlate with human experimental data as well as K\&C's results implied, and indeed not as well as that of A\&H's much earlier rule-based model. 

One issue in particular seems to be over-production of irregulars, which the model consistently prefers to regulars for four verbs (7\% of considered nonce verbs), while humans nearly always prefer the regular form. This was an issue with earlier neural net models as well \citep{plunkettjuola}. On the other hand, when the model outputs something other than the regular form, its choices are plausible. This was not true for earlier models: \citeauthor{plunkettjuola}'s model often chose the wrong regular suffix (with incorrect voicing in the final phoneme), and \citeauthor{rumelhartmcclelland}'s (\citeyear{rumelhartmcclelland}) model failed to produce regular endings for nonce verbs \cite{prasada1993,marcus1998}.
Here, we see from both our model's output and its internal representations that it has correctly identified the necessary voicing distinctions and that nonce words trigger similar representations and behaviour to real words. In future, a stricter test might use nonce words that are intentionally less similar to real words (e.g., the example from \citet{prasada1993}: \emph{to out-Gorbachev}).

It is also worth pointing out that the ED model, unlike A\&H's model and many earlier connectionist models, 
is fed raw phonemes (rather than the phonemes' distinctive features) as input. Although it learns some of the relevant features anyway, it would be interesting to see whether its behaviour becomes more human-like if the correct features are provided in the input.

Although our paper has revealed a number of weaknesses of the ED model, we do agree with K\&C that neural network-based cognitive models of inflection deserve re-evaluation in light of recent technical advances. There are many other potential architectures and modelling decisions that could be explored, as well as other behavioural data such as developmental patterns \citep{blything2018, ambridge2010} and inflection in other languages \citep[e.g.,][]{clahsen1992, ernestus2004}. As noted by \citet{seidenberg2014}, models' failures as well as successes can be informative, and we hope that our detailed exploration of the ED model's behaviour will inspire future developments in these models, both for cognitive modelling and NLP.

\section*{Acknowledgements}
This work was supported in part by a James S McDonnell Foundation Scholar Award (220020374).

\bibliography{thesis_refs}

\begin{thebibliography}{30}
\expandafter\ifx\csname natexlab\endcsname\relax\def\natexlab#1{#1}\fi

\bibitem[{Albright and Hayes(2003)}]{albrighthayes}
Adam Albright and Bruce Hayes. 2003.
\newblock \href {https://doi.org/10.1016/s0010-0277(03)00146-x} {Rules vs.
  analogy in {English} past tenses: a computational/experimental study}.
\newblock \emph{Cognition}, 90:119--161.

\bibitem[{Ambridge(2010)}]{ambridge2010}
Ben Ambridge. 2010.
\newblock \href {https://doi.org/10.1037/a0020668} {Children's judgments of
  regular and irregular novel past-tense forms: New data on the {English}
  past-tense debate.}
\newblock \emph{Developmental Psychology}, 46:1497--1504.

\bibitem[{Baayen et~al.(1995)Baayen, Piepenbrock, and Gulikers}]{celex}
R.~Harald Baayen, Richard Piepenbrock, and Leon Gulikers. 1995.
\newblock {CELEX2 LDC96L14. Web Download}.
\newblock Linguistic Data Consortium, Philadelphia, PA.

\bibitem[{Bahdanau et~al.(2015)Bahdanau, Cho, and Bengio}]{bahdanau}
Dzmitry Bahdanau, Kyunghyun Cho, and Yoshua Bengio. 2015.
\newblock \href {http://arxiv.org/abs/1409.0473} {Neural machine translation by
  jointly learning to align and translate}.
\newblock In \emph{3rd International Conference on Learning Representations,
  {ICLR} 2015}, San Diego, CA, USA.

\bibitem[{Bechtel and Abrahamsen(1991)}]{bechtel1991}
William Bechtel and Adele Abrahamsen. 1991.
\newblock \emph{Connectionism and the mind: An introduction to parallel
  processing in networks}.
\newblock Basil Blackwell, Oxford, England.

\bibitem[{Bergmanis and Goldwater(2018)}]{bergmanis2018}
Toms Bergmanis and Sharon Goldwater. 2018.
\newblock \href {https://doi.org/10.18653/v1/n18-1126} {Context sensitive
  neural lemmatization with {Lematus}}.
\newblock In \emph{Proceedings of the 2018 Conference of the {North American}
  Chapter of the {Association for Computational Linguistics}: Human Language
  Technologies, Volume 1 (Long Papers)}, pages 1391--1400. Association for
  Computational Linguistics.

\bibitem[{Blything et~al.(2018)Blything, Ambridge, and Lieven}]{blything2018}
Ryan~P. Blything, Ben Ambridge, and Elena~V.M. Lieven. 2018.
\newblock \href {https://doi.org/10.1111/cogs.12581} {Children's acquisition of
  the english past-tense: Evidence for a single-route account from novel verb
  production data}.
\newblock \emph{Cognitive Science}, 42:621--639.

\bibitem[{Bybee and Thompson(1997)}]{bybee1997}
Joan Bybee and Sandra Thompson. 1997.
\newblock \href {https://doi.org/10.3765/bls.v23i1.1293} {Three frequency
  effects in syntax}.
\newblock In \emph{Proceedings of the 23rd Annual Meeting of the Berkeley
  Linguistics Society}, pages 378--388. Berkeley Linguistics Society, Berkeley,
  CA.

\bibitem[{Clahsen et~al.(1992)Clahsen, Rothweiler, Woest, and
  Marcus}]{clahsen1992}
Harald Clahsen, Monika Rothweiler, Andreas Woest, and Gary~F. Marcus. 1992.
\newblock \href {https://doi.org/10.1016/0010-0277(92)90018-d} {Regular and
  irregular inflection in the acquisition of {German} noun plurals}.
\newblock \emph{Cognition}, 45:225--255.

\bibitem[{Elman et~al.(1996)Elman, Bates, Johnson, Karmiloff-Smith, Parisi, and
  Plunkett}]{elman96}
Jeffrey Elman, Elizabeth Bates, Mark~H. Johnson, Anette Karmiloff-Smith,
  Domenico Parisi, and Kim Plunkett. 1996.
\newblock \href {https://doi.org/10.7551/mitpress/5929.001.0001}
  {\emph{Rethinking innateness: {A} connectionist perspective on development}}.
\newblock MIT Press, Cambridge, MA.

\bibitem[{Ernestus and Baayen(2004)}]{ernestus2004}
Mirjam Ernestus and R.~Harald Baayen. 2004.
\newblock \href {https://doi.org/10.1515/ling.2004.031} {Analogical effects in
  regular past tense production in {Dutch}}.
\newblock \emph{Linguistics}, 42:873--903.

\bibitem[{Goldwater et~al.(2006)Goldwater, Griffiths, and
  Johnson}]{goldwater2006}
Sharon Goldwater, Thomas~L. Griffiths, and Mark Johnson. 2006.
\newblock \href
  {http://papers.nips.cc/paper/2941-interpolating-between-types-and-tokens-by-estimating-power-law-generators}
  {Interpolating between types and tokens by estimating power-law generators}.
\newblock In \emph{Advances in NIPS-18}, pages 459--466. Curran Associates,
  Inc., Red Hook, NY.

\bibitem[{Kann and Sch{\"u}tze(2016)}]{kann2016}
Katharina Kann and Hinrich Sch{\"u}tze. 2016.
\newblock \href {https://doi.org/10.18653/v1/w16-2010} {{MED}: The {LMU} system
  for the {SIGMORPHON} 2016 shared task on morphological reinflection}.
\newblock In \emph{{Proceedings of the 14th Annual SIGMORPHON Workshop on
  Computational Research in Phonetics, Phonology, and Morphology}}, pages
  62--70. Association for Computational Linguistics, Stroudsburg, PA.

\bibitem[{Kirov and Cotterell(2018)}]{kirovcotterell}
Christo Kirov and Ryan Cotterell. 2018.
\newblock \href {https://doi.org/10.1162/tacl_a_00247} {Recurrent neural
  networks in linguistic theory: Revisiting {Pinker and Prince} (1988) and the
  past tense debate}.
\newblock \emph{Transactions of the Association for Computational Linguistics},
  6:651--665.

\bibitem[{Klein et~al.(2017)Klein, Kim, Deng, Senellart, and Rush}]{opennmt}
Guillaume Klein, Yoon Kim, Yuntian Deng, Jean Senellart, and Alexander Rush.
  2017.
\newblock \href {https://doi.org/10.18653/v1/p17-4012} {{OpenNMT}: Open-source
  toolkit for neural machine translation}.
\newblock In \emph{Proceedings of the 55th Annual Meeting of the Association
  for Computational Linguistics, {ACL} 2017, System Demonstrations}, pages
  67--72. Association for Computational Linguistics.

\bibitem[{Linzen(2019)}]{linzen2019}
Tal Linzen. 2019.
\newblock \href {https://doi.org/10.1353/lan.2019.0001} {{What can linguistics
  and deep learning contribute to each other? Response to Pater}}.
\newblock \emph{Language}, 95:99--108.

\bibitem[{Linzen and Leonard(2018)}]{linzen2018}
Tal Linzen and Brian Leonard. 2018.
\newblock \href {https://mindmodeling.org/cogsci2018/papers/0147/index.html}
  {Distinct patterns of syntactic agreement errors in recurrent networks and
  humans}.
\newblock In \emph{{Proceedings of the 40th Annual Conference of the Cognitive
  Science Society}}, pages 692--697. Cognitive Science Society, Austin, TX.

\bibitem[{Marcus(1995)}]{marcus1995}
Gary~F. Marcus. 1995.
\newblock \href {https://doi.org/10.1016/0010-0277(94)00656-6} {{The
  acquisition of the English past tense in children and multilayered
  connectionist networks}}.
\newblock \emph{Cognition}, 56:271--279.

\bibitem[{Marcus(1998)}]{marcus1998}
Gary~F. Marcus. 1998.
\newblock \href {https://doi.org/10.1016/s0010-0277(98)00018-3} {Can
  connectionism save constructivism?}
\newblock \emph{Cognition}, 66:153--182.

\bibitem[{McCloskey(1991)}]{mccloskey1991}
Michael McCloskey. 1991.
\newblock \href {https://doi.org/10.1111/j.1467-9280.1991.tb00173.x} {Networks
  and theories: The place of connectionism in cognitive science}.
\newblock \emph{Psychological Science}, 2:387--395.

\bibitem[{O'Donnell(2015)}]{odonnell2015}
Timothy~J. O'Donnell. 2015.
\newblock \href {https://doi.org/10.7551/mitpress/9780262028844.001.0001}
  {\emph{Productivity and reuse in language: A theory of linguistic computation
  and storage}}.
\newblock MIT Press, Cambridge, MA.

\bibitem[{Pierrehumbert(2001)}]{pierrehumbert2001}
Janet Pierrehumbert. 2001.
\newblock \href {https://doi.org/10.1.1.132.5554} {Stochastic phonology}.
\newblock \emph{Glot International}, 5:195--207.

\bibitem[{Pinker and Prince(1988)}]{pinkerprince}
Steven Pinker and Alan Prince. 1988.
\newblock \href {https://doi.org/10.1016/0010-0277(88)90032-7} {On language and
  connectionism: Analysis of a parallel distributed processing model of
  language acquisition}.
\newblock \emph{Cognition}, 28:73--193.

\bibitem[{Pinker and Ullman(2002)}]{pinker2002}
Steven Pinker and Michael~T. Ullman. 2002.
\newblock \href {https://doi.org/10.1016/s1364-6613(02)01990-3} {The past and
  future of the past tense}.
\newblock \emph{Trends in Cognitive Sciences}, 6:456--463.

\bibitem[{Plunkett and Juola(1999)}]{plunkettjuola}
Kim Plunkett and Patrick Juola. 1999.
\newblock \href {https://doi.org/10.1207/s15516709cog2304_4} {A connectionist
  model of {English} past tense and plural morphology}.
\newblock \emph{Cognitive Science}, 23:463--490.

\bibitem[{Prasada and Pinker(1993)}]{prasada1993}
Sandeep Prasada and Steven Pinker. 1993.
\newblock \href {https://doi.org/10.1080/01690969308406948} {Generalisation of
  regular and irregular morphological patterns}.
\newblock \emph{Language and Cognitive Processes}, 8:1--56.

\bibitem[{Rumelhart and McClelland(1986)}]{rumelhartmcclelland}
David~E. Rumelhart and James~L. McClelland. 1986.
\newblock On learning the past tenses of {English} verbs.
\newblock In James~L. McClelland, David~E. Rumelhart, and the PDP
  Research~Group, editors, \emph{Parallel distributed processing: Explorations
  in the microstructure of cognition}, chapter~18, pages 216--271. MIT Press,
  Cambridge, MA.

\bibitem[{Seidenberg and Plaut(2014)}]{seidenberg2014}
Mark~S. Seidenberg and David~C. Plaut. 2014.
\newblock \href {https://doi.org/10.1111/cogs.12147} {Quasiregularity and its
  discontents: The legacy of the past tense debate}.
\newblock \emph{Cognitive Science}, 38:1190--1228.

\bibitem[{Sutskever et~al.(2014)Sutskever, Vinyals, and Le}]{sutskever2014}
Ilya Sutskever, Oriol Vinyals, and Quoc~V. Le. 2014.
\newblock \href
  {http://papers.nips.cc/paper/5346-sequence-to-sequence-learning-with-neural-networks}
  {Sequence to sequence learning with neural networks}.
\newblock In \emph{Advances in Neural Information Processing Systems 27: Annual
  Conference on Neural Information Processing Systems 2014}, pages 3104--3112.
  Curran Associates, Inc., Red Hook, NY.

\bibitem[{Zeiler(2012)}]{zeiler}
Matthew~D. Zeiler. 2012.
\newblock \href {http://arxiv.org/abs/1212.5701} {{ADADELTA}: An adaptive
  learning rate method}.
\newblock \emph{Computing Research Repository}, arXiv:1212.5701.

\end{thebibliography}
\bibliographystyle{acl_natbib}

\end{document}